%% file: paper.tex
\documentclass{article}
\usepackage{graphicx} 
\usepackage{subfigure} 
\usepackage{natbib}
\usepackage{algorithm}
\usepackage{algorithmic}
\usepackage{afterpage}
\usepackage{enumitem}

\usepackage{hyperref}

\usepackage[accepted]{icml2012}

\usepackage{tikz}
\usepackage{pgfplots}
\pgfrealjobname{paper}

\tikzset{every picture/.style={font=\small}}

\input{stat-macros.tex}

\input{editing-macros.tex}
\newcommand{\rmax}{R-{\sc max}}

\input{trajectory_plot.tex}
\input{expl_success_plot.tex}

\input{running_time_plot.tex}
\input{counter_mdps.tex}
\input{np_hard_proof_diagram.tex}

\icmltitlerunning{Safe Exploration in Markov Decision Processes}
\newcommand{\ucberkeleyaddress}{University of California at Berkeley, CA 94720-1758, USA}

\begin{document} 
\twocolumn[
\icmltitle{Safe Exploration in Markov Decision Processes}

\icmlauthor{Teodor Mihai Moldovan}{moldovan@cs.berkeley.edu}
\icmlauthor{Pieter Abbeel}{pabbeel@cs.berkeley.edu}
\icmladdress{\ucberkeleyaddress}

\icmlkeywords{machine learning, ICML, safety, exploration, safe exploration, markov decision process, MDP, grid world, np-hard, approximate, approximation, Mars, High Resolution Imaging Science Experiment, HiRISE, Mars Science Laboratory, MSL}

\vskip 0.3in
]
\begin{abstract}
In environments with uncertain dynamics exploration is necessary to learn how to perform well. Existing reinforcement learning algorithms provide strong exploration guarantees, but they tend to rely on an ergodicity assumption. The essence of ergodicity is that any state is eventually reachable from any other state by following a suitable policy. This assumption allows for exploration algorithms that operate by simply favoring states that have rarely been visited before. For most physical systems this assumption is impractical as the systems would break before any reasonable exploration has taken place, i.e., most physical systems don't satisfy the ergodicity assumption. In this paper we address the need for \emph{safe} exploration methods in Markov decision processes. We first propose a general formulation of safety through ergodicity.   We show that imposing safety by restricting attention to the resulting set of guaranteed safe policies is NP-hard.   We then present an efficient algorithm for guaranteed safe, but potentially suboptimal, exploration.  At the core is an optimization formulation in which the constraints restrict attention to a subset of the guaranteed safe policies and the objective favors exploration policies. Our framework is compatible with the majority of previously proposed exploration methods, which rely on an exploration bonus. Our experiments, which include a Martian terrain exploration problem, show that our method is able to explore better than classical exploration methods.
\end{abstract}

\input{introduction}
\input{background}
\input{objective}
\input{algorithm}
\input{experiments}

\input{discussion}


{\small\bibliography{library}}
\bibliographystyle{icml2012}

\newpage
\input{appendix}

\end{document}

%% file: stat-macros.tex
\usepackage{amsmath,amssymb,amsthm}
\input{math-macros}
\input{thm-macros}

%% file: math-macros.tex
\usepackage{amsmath}

\newcommand{\ind}[1]{1_{#1}} 
\newcommand{\ex}{E} 
\newcommand{\var}{\textrm{Var}}

\renewcommand{\[}{\begin{equation}}
\renewcommand{\]}{\end{equation}}
\renewcommand{\b}{\backslash}

\newcommand{\bea}{\begin{eqnarray}}
\newcommand{\eea}{\end{eqnarray}}

\newcommand{\deq}{:=}

\newcommand{\argmax}{\mathop{\mathrm{argmax}}}


%% file: thm-macros.tex
\theoremstyle{plain}
\newtheorem{theorem}{Theorem}

\newtheorem{lemma}[theorem]{Lemma}

\theoremstyle{definition}

\theoremstyle{remark}

%% file: trajectory_plot.tex
\colorlet{backgroundcolor}{gray!15!white}
\colorlet{colormap5}{violet}
\colorlet{colormap4}{red}
\colorlet{colormap3}{brown}
\colorlet{colormap2}{green!50!black}
\colorlet{colormap1}{blue}
\newcommand{\plotpath}[2]{\def\filename{#1}\beginpgfgraphicnamed{./figures/#2}\def\xlen{3ex}\def\ylen{\xlen}\def\rotation{-90}\begin{tikzpicture}[inner sep = 0, rotate = \rotation, >= to, shorten <= 0 ex, x=\ylen, y=\xlen,] 

\input{\filename}

\
\tikzstyle{sq}=[minimum width = \xlen, minimum height = \ylen]

\foreach \x / \y / \z / \d  in \mapdata
{
\node[sq,
draw=backgroundcolor, 
fill=backgroundcolor,
] at (\x,\y)  {};
\ifnum \d=1
\edef\currentcolor{colormap\z!30!white}
\node[rounded corners = .2\xlen,sq,
draw=\currentcolor!100!backgroundcolor, 
fill=\currentcolor!100!backgroundcolor,
minimum width = \xlen * (.4 + \z*.1),
minimum height = \ylen* ( .4 + \z*.1),
] at (\x,\y)  {};
\else
\edef\currentcolor{backgroundcolor}
\node[sq,
draw=\currentcolor!50!backgroundcolor, 
fill=\currentcolor,
] at (\x,\y)  {};
\fi
}

\tikzstyle{pth}=[ thick, |-, > = stealth, opaque, rounded corners=\xlen*.1]
\draw[pth]
\foreach \x / \y [remember=\x as \lastpos] in \pathdata
{
\ifx\lastpos\undefined\else--\fi(\x,\y)
};

\tikzstyle{arrowstyle}=[  pth,  -> , shorten >= \xlen*.0]
\foreach \a / \b / \c / \d in \arrowdata
{
\draw[arrowstyle] (\a,\b)--(\c,\d);
}

\end{tikzpicture}
\endpgfgraphicnamed
}

%% file: expl_success_plot.tex
\newcommand{\plotexplsucc}[1]{\beginpgfgraphicnamed{./figures/#1}\begin{tikzpicture}
\pgfplotsset{/pgfplots/error bars/error mark options={
      rotate=90,
      mark size=3pt,
    }}

\pgfplotsset{every x tick/.append style={transparent}}

\begin{axis}[
	ylabel=Fraction of grid world uncovered,
	xlabel=\raisebox{-12pt}{Grid world size $|$ Mean fraction of inaccessible squares (walls)},
	ymin=0, ymax=1.05,
	xmin=.5,xmax=12.5,
	legend style={at={(.5, 1.05)},
		anchor=south,legend columns=-1},
	legend style={/tikz/every even column/.append style={column sep=2ex}},
	ybar,
	bar width=6pt,
	width = \textwidth,
	height=.8\columnwidth,
	xtick=\empty,
	xtick={1,2,3,4,5,6,7,8,9,10,11,12},
	xticklabels={ 
		$10^2{|}.2$ ,
		$10^2{|}.4$ ,
		$10^2{|}.6$ ,
		$20^2{|}.2$ ,
		$20^2{|}.4$ ,
		$20^2{|}.6$ ,
		$30^2{|}.2$ ,
		$30^2{|}.4$ ,
		$30^2{|}.6$ ,
		$40^2{|}.2$ ,
		$40^2{|}.4$ ,
		$40^2{|}.6$ },
]

\foreach \file in {
	./data/expl_succ_safe_adapted_rmax.dat, 
	./data/expl_succ_adapted_rmax.dat,
	./data/expl_succ_rmax.dat%
	}
{
\addplot+[draw=none] table[forget plot,x=x,y=median] 
	{\file};

\addplot+[draw=none,fill=none] table[forget plot,x=x,y=median,y error=uqminusmedian, /pgfplots/error bars/.cd, y dir=plus,y explicit] 
	{\file};
\addplot+[draw=none,fill=none] table[forget plot,x=x,y=median,y error=medianminuslq, /pgfplots/error bars/.cd,
y dir=minus,y explicit] 
	{\file};

\addplot+[draw=none,fill=none,] table[forget plot,x=x,y=median, 
/pgfplots/error bars/.cd,
	y dir=minus,
	y fixed=0,] 
	{\file};

\addplot+[] coordinates
{(1,0) (2,0) (3,0) (4,0) (5,0) (6,0) (7,0) (8,0) (9,0)}; 

}

\legend{
~\shortstack{Safe version of adapted\\ \rmax~exploration},
~\shortstack{Adapted \rmax~\\exploration},
~\shortstack{Original \rmax~or\\ Near-Bayesian exploration}}
\end{axis}
\end{tikzpicture}
\endpgfgraphicnamed
}

%% file: counter_mdps.tex
\newcommand{\countermdpa}[1]{\beginpgfgraphicnamed{./figures/#1}
\begin{tikzpicture}[ 
state/.style={draw, circle},
node distance = 13ex,
auto,
->, >= stealth, semithick, shorten >= .2 ex,
] 

\node(B) [state] {B};
\node(S) [state,left of= B] {S};
\node(F) [state,above right of= B] {F};
\node(E) [state,above right of= S] {E};

\draw[->] (S) to[bend left] node{a .2} node[swap]{} (E);
\draw[->] (S) to[] node[swap]{a .8}  (B);
\draw[->] (B) to[bend right] node[swap]{b}  (S);
\draw[->] (B) to[loop right] node{c .9}  (B);
\draw[->] (B) to[bend left] node{a}  (F);
\draw[->] (F) to[] node{b}  (B);
\draw[->] (B) to[bend left] node[swap]{c .1}  (E);

\end{tikzpicture}
\endpgfgraphicnamed
}
\newcommand{\countermdpb}[1]{\beginpgfgraphicnamed{./figures/#1}
\begin{tikzpicture}[ 
state/.style={draw, circle},
node distance = 8ex,
auto,
->, >= stealth, semithick, shorten >= .2 ex,
] 

\node(L1) [state] {L};
\node(B1) [state, left of=L1] {B};
\node(U1) [state, left of=B1] {U};
\node(S1) [state, above of=B1] {S};

\node(U2) [state, right of=L1] {U};
\node(B2) [state, right of=U2] {B};
\node(L2) [state, right of=B2] {L};
\node(S2) [state, above of=B2] {S};

\draw[->] (S1) to node{a}  (B1);
\draw[->] (S1) to[loop right] node{b}  (S1);
\draw[->] (B1) to node{a}  (U1);
\draw[->] (B1) to node{b}  (L1);
\draw[->] (L1) to[loop above] node{a}  (L1);
\draw[->] (U1) to[bend left] node{a}  (S1);

\draw[->] (S2) to node{a}  (B2);
\draw[->] (S2) to[loop left] node{b}  (S2);
\draw[->] (B2) to node{b}  (L2);
\draw[->] (B2) to node{a}  (U2);
\draw[->] (U2) to[loop above] node{a}  (U2);
\draw[->] (L2) to[bend right] node[swap]{a}  (S2);

\end{tikzpicture}
\endpgfgraphicnamed
}
\newcommand{\countermdpc}[1]{\beginpgfgraphicnamed{./figures/#1}
\begin{tikzpicture}[ 
state/.style={draw, circle},
node distance = 9ex,
auto,
->, >= stealth, semithick, shorten >= .2 ex,
] 

\node(A1) [state] {A};
\node(C1) [state, right of=A1] {C};
\node(B1) [state, above left of=C1] {B};

\node(A2) [state, right of=C1] {A};
\node(C2) [state, right of=A2] {C};
\node(B2) [state, above left of=C2] {B};

\node(A3) [state, right of=C2] {A};
\node(C3) [state, right of=A3] {C};
\node(B3) [state, above left of=C3] {B};

\draw[->] (A1) to[] node{a}  (B1);
\draw[->] (B1) to[bend right] node{a}  (C1);
\draw[->] (C1) to[loop above] node{a}  (C1);

\draw[->] (A2) to[] node{a}  (B2);
\draw[->] (B2) to[loop left] node{a}  (B2);
\draw[->] (C2) to[] node{a}  (A2);

\draw[->] (B3) to[bend left] node[swap]{a .5}  (C3);
\draw[->] (B3) to[loop left] node{a .5}  (B3);
\draw[->] (C3) to[] node{a .5}  (A3);
\draw[->] (C3) to[loop above] node{a .5}  (C3);
\draw[->] (A3) to[] node{a}  (B3);

\end{tikzpicture}
\endpgfgraphicnamed
}

%% file: np_hard_proof_diagram.tex
\newcommand{\nphardproofdiagram}[1]{\beginpgfgraphicnamed{./figures/#1}
\begin{tikzpicture}[ 
state/.style={draw, circle},
node distance = 8ex, minimum size = 5ex,
auto,
->, >= stealth, semithick, shorten >= .2 ex,
] 
\node(X1) [state] {$X_{1}$};
\node(Y1) [state, below of=X1] {$Y_{1}$};
\node(Z1) [state, below of=Y1] {$Z_{1}$};

\node(X2) [state, right of=X1]{$X_{2}$};
\node(Y2) [state, below of=X2] {$Y_{2}$};
\node(Z2) [state, below of=Y2] {$Z_{2}$};

\node(Xi) [right of=X2] {\ldots};
\node(Yi) [below of=Xi] {\ldots};
\node(Zi) [below of=Yi] {\ldots};

\node(Xn) [state, right of=Xi]{$X_{n}$};
\node(Yn) [state, below of=Xn] {$Y_{n}$};
\node(Zn) [state, below of=Yn] {$Z_{n}$};

\node(S) [state, right of=Yn] {D};
\node(S2) [state, left of=Y1] {S};

\draw[->] (S) to[bend left,in=90,out=90]  (S2);

\draw[->] (X1) to (X2);
\draw[->] (X1) to (Y2);
\draw[->] (X1) to (Z2);
\draw[->] (Y1) to (X2);
\draw[->] (Y1) to (Y2);
\draw[->] (Y1) to (Z2);
\draw[->] (Z1) to (X2);
\draw[->] (Z1) to (Y2);
\draw[->] (Z1) to (Z2);

\draw[->] (X2) to (Xi);
\draw[->] (X2) to (Yi);
\draw[->] (X2) to (Zi);
\draw[->] (Y2) to (Xi);
\draw[->] (Y2) to (Yi);
\draw[->] (Y2) to (Zi);
\draw[->] (Z2) to (Xi);
\draw[->] (Z2) to (Yi);
\draw[->] (Z2) to (Zi);

\draw[->] (Xi) to (Xn);
\draw[->] (Xi) to (Yn);
\draw[->] (Xi) to (Zn);
\draw[->] (Yi) to (Xn);
\draw[->] (Yi) to (Yn);
\draw[->] (Yi) to (Zn);
\draw[->] (Zi) to (Xn);
\draw[->] (Zi) to (Yn);
\draw[->] (Zi) to (Zn);

\draw[->] (Xn) to (S);
\draw[->] (Yn) to (S);
\draw[->] (Zn) to (S);

\draw[->] (S2) to (X1);
\draw[->] (S2) to (Y1);
\draw[->] (S2) to (Z1);

\end{tikzpicture}
\endpgfgraphicnamed
}

%% file: introduction.tex
\section{Introduction}
\label{sec:intro}

When humans learn to control a system, they naturally account for what we think of as safety. For example, when a novice pilot learns how to fly an RC helicopter, they will slowly spin up the blades until the helicopter barely lifts off, then quickly put it back down. They will repeat this a few times, slowly starting to bring the helicopter a little bit off the ground. When doing so they would try out the cyclic (roll and pitch) and rudder (yaw) control, while---until they have become more skilled---at all times staying low enough that simply shutting it down would still have it land safely. When a driver wants to become skilled at driving on snow, they might first slowly drive the car to a wide open space where they could start pushing their limits. When we are skiing downhill, we are careful about not going down a slope into a valley where there is no lift to take us back up.

One would hope that exploration algorithms for physical systems would be able to account for safety and have similar behavior naturally emerge.  Unfortunately most existing exploration algorithms completely ignore safety issues.  More precisely phrased, most existing algorithms have strong exploration guarantees, but to achieve these guarantees they assume ergodicity of the Markov decision process (MDP) in which the exploration takes place.   An MDP is \emph{ergodic} if any state is reachable from any other state by following a suitable policy.  This assumption does not hold true in the exploration examples presented above as each of these systems could break during (non-safe) exploration.  

Our first important contribution is a definition of safety, which, at its core, requires restricting attention to policies that preserve ergodicity with some well controlled probability. Imposing safety is, unfortunately, NP-hard in general. Our second important contribution is an approximation scheme leading to guaranteed safe, but potentially sub-optimal, exploration.\footnote{Note that existing (unsafe) exploration algorithms are also sub-optimal, in that they are not guaranteed to complete exploration in the minimal number of time steps.} A third contribution is the consideration of uncertainty in the dynamics model that is correlated over states. While usually the assumption is that uncertainty in different parameters is independent---as this makes problem more tractable computationally---being able to learn about state-action pairs before visiting them is critical for safety.

Our experiments illustrate that our method indeed achieves safe exploration, in contrast to plain exploration methods. They also show that our algorithm is almost as computationally efficient as planning in a known MDP---but then, as every step leads to an update in knowledge about the MDP, this computation is to be repeated after every step. Our approach is able to safely explore grid worlds of size up to 50 × 100. Our method can make safe any type of exploration that relies on exploration bonuses, which is the case for most existing exploration algorithms, including, for example, the methods proposed in \cite{Brafman2001,Kolter2009}. In this article we do not focus on the exploration objective and use existing ones.



Safe exploration has been the focus of a large number of articles. \cite{Gillula2011,Aswani2012a} propose safe exploration methods for linear systems with bounded disturbances based on model predictive control and reachability analysis. They define safety in terms of safe regions of the state space, which, we will show, is not always appropriate in the context of MDPs. The safe exploration for MDP methods proposed by \cite{Geramifard11ACC,Hans2008} gauge safety based on the best best estimate of the transition measure but they ignore the level of uncertainty in this estimate. As we will show, this is not sufficient to provably guarantee safety. 

Provably efficient exploration is a recurring theme in reinforcement learning \cite{Strehl2005,Li2008,Brafman2001,Kearns2002,Kolter2009}. Most methods, however, tend to rely on the assumption of ergodicity which rarely holds in interesting practical examples; consequently, these methods are rarely applicable for physical systems. The issue of provably guaranteed safety, or risk aversion, under uncertainty in the MDP parameters has also been studied in the reinforcement literature.  In \cite{Nilim2005} they propose a robust MDP control method assuming the transition frequencies are drawn from an orthogonal convex set by an adversary. Unfortunately, it seems impossible to use their method to constrain some safety objective while optimizing a different exploration objective. In \cite{Delage2007} they present a safe exploration algorithm for the special case of Gaussian distributed ambiguity in the reward and state-action-state transition probabilities, but their safety guarantees are only accurate if the ambiguity in the transition model is small.

%% file: background.tex
\section{Notation and Assumptions}
\label{sec:background}
Due to space constraints, we will not give a general introduction to Markov decision processes (MDPs).  For an introduction to MDPs we refer the readers to \cite{Sutton1998,Bertsekas1996}. 

We use capital letters to denote random variables; for example, the total reward is: $V \deq \sum_{t=0}^\infty R_{S_t, A_t}$. We represent the policies and the initial state distributions by probability measures. Usually the measure $\pi$ will correspond to a policy and the measure $s \deq \delta(s)$, which puts measure only in state $s$, will correspond to starting in state $s$. With this notation, the usual value recursion, assuming a known transition measure, $p$, reads:
\begin{align*}
  \ex_{s,\pi}^p [V] = \sum_{a,s'} \pi_{s,a} \left( \ex[R]_{s,a} + p_{s,a,s'} \ex_{s',\pi}^p [V]\right).
\end{align*}
We specify the transition measure as a superscript of the expectation operator rather than a subscript for typographical convenience; in this case, and in general, the positioning of indexes as subscripts or superscripts adds no extra significance.  We will let the transition measure $p$ sometimes sum to less than one, that is $\sum_{s'} p_{s,a,s'} \leq 1$. The missing mass is implicitly assigned to transitioning to an absorbing ``end'' state, which, for example, allows us to model $\gamma$ discounting by simply using $\gamma p$ as a transition measure.

We model ambiguous dynamics in a Bayesian way, allowing the transition measure to also be a random variable. When this is the case, we will use $P$ to denote the, now random, transition measure. The \emph{belief}, which we will denote by $\beta$, is our Bayesian probability measure over possible dynamics, governing $P$ and $R$. Therefore, the expected return under the belief and policy $\pi$, starting from state $s$, is $\ex_{\beta} \ex_{s,\pi}^P[V]$. We allow beliefs under which transition measures and rewards are arbitrarily correlated. In fact, such correlations are usually necessary to allow for safe exploration.
For compactness we will often use lower case letters to denote the expectation of their upper case counterparts. Specifically, we will use the notations $p\deq \ex_{\beta}[P]$ and $r \deq \ex_{\beta}[R]$ throughout. 

%% file: objective.tex
\section{Problem formulation}
\label{sec:objective}
\subsection{Exploration Objective}
Exploration methods, as those proposed in \cite{Brafman2001,Kolter2009}, operate by finding optimal policies in constructed MDPs with exploration bonuses. The \rmax~algorithm, for example, constructs an MDP based on the discounted expected transition measure and rewards under the belief, and adds a deterministic exploration bonus equal to the maximum possible reward in the MDP, $\xi^\beta_{s,a} = r_{\text{max}}$, to any transitions that are not sufficiently well known. Our method allows adding safety constraints to any such exploration methods. Henceforth, we will restrict attention to such exploration methods, which can be formalized as optimization problems of the form:
\begin{align}
\label{eq:expl_obj}
&\text{maximize }_{\pi_o}~ \ex_{s_0, \pi_o}^{\gamma p} \sum_{t=0}^{\infty} \left(r_{S_t, A_t} + \xi^\beta_{S_t, A_t} \right).
\end{align}
\subsection{Safety Constraint}
\label{sec:safety-constraint}
The issue of safety is closely related to \emph{ergodicity}. Almost all proposed exploration techniques presume ergodicity; authors present it as a harmless technical assumption but it rarely holds in interesting practical problems. Whenever this happens, their efficient exploration guarantees cease to hold, often leading to very inefficient policies. Informally, an environment is ergodic if any mistake can be forgiven eventually. More specifically, a belief over MDPs is ergodic if and only if any state is reachable from any other state via some policy or, equivalently, if and only if: 
\begin{align}
\label{eq:ergodicity}
\forall s,s', \exists~\pi_r \text{ such that } \ex_{\beta} \ex_{s,\pi_r}^P [B_{s'} ]=1,
\end{align} where $B_{s'}$ is an indicator random variable of the event that the system reaches state $s'$ at least once: $B_{s'} = \ind{}\{ \exists t<\infty \text{ such that } S_t=s' \} = \min\left(1, \sum_{t} \ind{S_{t} =s'}\right)$.

Unfortunately, many environments are not ergodic. For example, our robot helicopter learning to fly cannot recover on its own after crashing. Ensuring almost sure ergodicity is too restrictive for most environments as, typically, there always is a very small, but non-zero, chance of encountering that particularly unlucky sequence of events that breaks the system. Our idea is to restrict the space of eligible policies to those that preserve ergodicity with some user-specified probability, $\delta$, called the \emph{safety level}. We name these policies \emph{$\delta$-safe}. Safe exploration now amounts to choosing the best exploration policy from this set of safe policies. 

Informally, if we stopped a $\delta$-safe policy $\pi_{o}$ at any time $T$, we would be able to return from that point to the home state $s_0$ with probability $\delta$ by deploying a return policy $\pi_r$. Executing only $\delta$-safe policies in the case of a robot helicopter learning to fly will guarantee that the helicopter is able to land safely with probability $\delta$ whenever we decide to end the experiment. In this example, $T$ is the time when the helicopter is recalled (perhaps because fuel is running low), so we will call $T$ the \emph{recall time}. Formally, an outbound policy $\pi_{o}$ is $\delta$-safe with respect to a home state $s_0$ and a stopping time $T$ if and only if:
\begin{align}
\label{eq:psafe}
\exists \pi_r \text{ such that } \ex_{\beta} \ex_{s_{0},\pi_{o}}^P \left[\ex_{S_T,\pi_{r}}^P [B_{s_{0}}]\right] \geq \delta.
\end{align}
Note that, based on Equation~\eqref{eq:ergodicity}, any policy is $\delta$-safe for any $\delta$ if the MDP is ergodic with probability one under the belief. For convenience we will assume that the recall time, $T$, is exponentially distributed with parameter $1-\gamma$, but our method also applies when the recall time equals some deterministic horizon. Unfortunately, expressing the set of $\delta$-safe policies is NP-hard in general, as implied by the following theorem proven in the appendix. 
\begin{theorem}\label{th:np-hard} In general, it is NP-hard to decide whether there exist $\delta$-safe policies with respect to a home state, $s_0$, and a stopping time, $T$, for some belief,~$\beta$.
\end{theorem}
\subsection{Safety Counter-Examples}
\label{sec:objective-counter}
\begin{figure}[t]
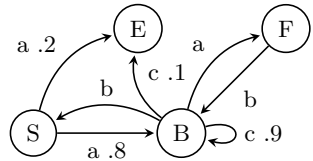

\centering
\countermdpa{counter_mdp_a}
\caption{\label{fig:counter-mdp1}Starting from state S, the policy (aababab\ldots) is safe at a safety level of $.8$. However, the policy (acccc\ldots) is not safe since it will end up in the sink state E with probability 1. State-action Sa and state B can neither be considered safe nor unsafe, since both policies use them.}
\end{figure}
We conclude this section with counter-examples to three other, perhaps at first sight more intuitive, definitions of safety. First, we could have tried to define safety in terms of sets of safe states or state-actions. That is, we might think that making the non-safe states and actions unavailable to the planner (or simply inaccessible) is enough to guarantee safety. Figure~\ref{fig:counter-mdp1} shows an MDP where the same state-action is used both by a safe and by an unsafe policy. The idea behind this counter-example is that safety depends not only on the states visited, but also on the number of visits, thus, on the policy. This shows that safety should be defined in terms of safe policies, not in terms of safe states or state-actions.

\begin{figure}[t]
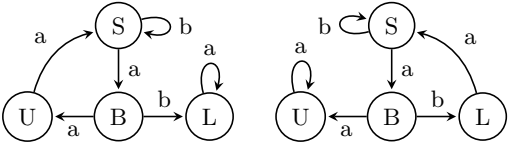

\centering
\countermdpb{counter_mdp_b}
\caption{\label{fig:counter-mdp2}Under out belief the two MDPs above both have probability $.5$. It is intuitively unsafe to go from the start state S to B since we wouldn't know whether the way back is via U or L, even though we know for sure that a return policy exists.}
\end{figure}
Second, we might think that it is perhaps enough to ensure that there exists a return policy for each potential sample MDP from the belief, but not impose that it be the same for all samples. That is, we might think that condition \ref{eq:psafe} is too strong and, instead, it would be enough to have: 
\begin{align*}
\ex_{\beta} \ind{}\{\exists \pi_{r} : \ex_{s_0,\pi_o}^P \ex_{S_T, \pi_{r}}^P [B_{s_0}] = 1 \} \geq \delta.
\end{align*}
Figure~\ref{fig:counter-mdp2} shows an MDP where this condition holds, yet all policies are naturally unsafe. 

\begin{figure}[t]
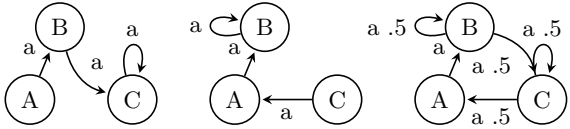

\centering
\countermdpc{counter_mdp_c}
\caption{\label{fig:counter-mdp3}The two MDPs on the left both have probability $.5$. Under this belief, starting from state A, policy (aaa\ldots) is unsafe. However, under the mean transition measure, represented by the MDP on the right, the policy is safe.} 
\end{figure}
Third, we might think that it is sufficient to simply use the expected transition measure when defining safety, as in the equation below. Figure~\ref{fig:counter-mdp3} shows that this is not the case; the expected transition measure is not a sufficient statistic for safety.
\begin{align*}
\exists \pi_r \text{ such that } \ex_{s_{0},\pi_{o}}^p \left[\ex_{S_T,\pi_{r}}^p [B_{s_{0}}]\right] \geq \delta.
\end{align*}

%% file: algorithm.tex
\section{Guaranteed Safe, Potentially Sub-optimal Exploration}
\begin{algorithm}[t]
\caption{Safe exploration algorithm}
\label{alg:safe_expl}
\begin{algorithmic}
\REQUIRE prior belief $\beta$, discount $\gamma$, safety level $\delta$.
\REQUIRE function $\xi:$ belief $\rightarrow$ exploration bonus
\STATE $M,N \leftarrow$ new MDP objects 
\REPEAT
\STATE $s_0, \varphi \leftarrow \text{current state and observations}$
\STATE update belief $\beta$ with information $\varphi$
\STATE $\xi_{s,a}^\beta \leftarrow \xi(\beta)$ (exploration bonus based on $\beta$)
\STATE $\sigma_{s,a}^\beta \leftarrow \sum_{s'} \ex_{\beta} [ \min(0, P_{s,a,s'}- \ex_\beta[P_{s,a,s'}])]$
\STATE $M.\text{transition measure} \leftarrow \ex_\beta[P] (1-\ind{s=s_0})$ 
\STATE $M.\text{reward function} \leftarrow \ind{s=s_0} + (1-\ind{s=s_0}) \sigma_{s,a}^\beta$ 
\STATE $\pi^r, v \leftarrow M.\text{solve()}$
\STATE $N.\text{transition measure} \leftarrow \gamma \ex_\beta[P] $ 
\STATE $N.\text{reward function} \leftarrow \ex_\beta[R_{s,a}] + \xi_{s,a}^\beta$
\STATE $N.\text{constraint reward func.} \leftarrow (1-\gamma) v_s + \gamma \sigma_{s,a}^\beta$
\STATE $N.\text{constraint lower bound} \leftarrow \delta$ 
\STATE $\pi^o, v^\xi, v^\sigma\leftarrow N.\text{solve under constraint()}$
\STATE $q^{\sigma}_{s,a} \leftarrow (1-\gamma) v_{s} + \gamma \sigma^\beta_{s,a} + \sum_{s'} p_{s,a,s'} v_{s'}^{\sigma}$
\STATE $a \leftarrow \argmax_{\{\pi^o_{s_0,a} > 0\}} q^\sigma_{s_0,a}$ (de-randomize policy)
\STATE take action $a$ in environment
\UNTIL{$\xi^\beta = 0$, so there is nothing left to explore}
\end{algorithmic}
\end{algorithm}
Although imposing the safety constraint in Equation~\eqref{eq:psafe} is NP-hard, as shown in Theorem~\ref{th:np-hard}, we can efficiently constrain a lower bound on the safety objective, so the safety condition is still provably satisfied. Doing so could lead to sub-optimal exploration since the set of policies we are optimizing over has shrunk. However, we should keep in mind that the exploration objectives represent approximate solutions to other NP-hard problems, so optimality has already been forfeited in existing (non-safe) approaches to start out with. Algorithm~\ref{alg:safe_expl} summarizes the procedure and the experiments presented in the next section show that, in practice, when the ergodicity assumptions are violated, safe exploration is much more efficient than plain exploration.

Putting together the exploration objective defined in Equation~\eqref{eq:expl_obj} and the safety objective defined in Equation~\eqref{eq:psafe} allows us to formulate safe exploration at level $\delta$ as a constrained optimization problem:
\begin{align*}
\text{maximize }_{\pi_o,\pi_r}\quad&\ex_{s_0, \pi_o}^{\gamma p} \sum_{t} \left(r_{S_t, A_t} + \xi^\beta_{S_t, A_t} \right) \\
\text{such that: }\quad&\ex_{\beta} \ex_{s_{0},\pi_o}^P \left[\ex_{S_T,\pi_{r}}^P [B_{s_{0}}]\right] \geq \delta.
\end{align*}
The exploration objective is already conveniently formulated as the expected reward in an MDP with transition measure $\gamma p$, so we will not modify it. On the other hand, the safety constraint is difficult to deal with as is. Ideally, we would like the safety constraint to also equal some expected reward in an MDP. We will see that, in fact, it takes two MDPs to express the safety constraint. 

First, we express the inner term, $\ex_{S_T,\pi_{r}}^P [B_{s_{0}}]$, as the expected reward in an MDP. We can replicate the behaviour of $B_{s_0}$, that is counting only the first time state $s_0$ is reached, by using a new transition measure, $P\cdot(1-\ind{s=s_0})$ under which, once $s_0$ is reached, any further actions lead immediately to the implicit ``end'' state. Formally, we express this by the identity:
\begin{align*}
\ex_{S_T,\pi_r}^P[B_{s_0}] = \ex_{S_T,\pi_r}^{P\cdot(1-\ind{s=s_0})} \sum_{t=0}^\infty  1_{S_t = s_0}.
\end{align*} 
We now focus on the outer term, $\ex_{s_0,\pi_o}^P \left[\ex_{S_T,\pi_{r}}^P [B_{s_{0}}]\right]$. Since the recall time, $T$, is exponentially distributed with parameter $1-\gamma$, we can view $S_T$ as the final state in a $\gamma$-discounted MDP starting at state $s_0$, following policy $\pi_o$. In this MDP, the inner term plays the role of a terminal reward. To put the problem in a standard form, we convert this terminal reward to a step-wise reward by multiplying it by $1-\gamma$.
\begin{align*}
\ex_{s_0,\pi_o}^P \left[\ex_{S_T,\pi_{r}}^P [B_{s_{0}}]\right] = \ex_{s_0,\pi_o}^{\gamma P} \sum_{t=0}^{\infty} (1-\gamma) \left[\ex_{S_t,\pi_{r}}^P [B_{s_{0}}]\right].
\end{align*}
At this point, we have expressed the safety constraint in the MDP formalism, but the transition measures of these MDPs, $P(1-\ind{s=s_0})$ and $\gamma P$, are still random.  If we could replace these random transition measures with their expectations under the belief $\beta$ that would significantly simplify the safety constraint. It turns out we can do this, at the expense of making the constraint more stringent. Our tool for doing so is Theorem~\ref{th:approx-correction} presented below, but proven in the appendix. It shows that we can replace a belief over MDPs by a single MDP with the expected transition measure, featuring an appropriate reward correction such that, under any policy, the value of this MDP is a lower bound on the expected value under the belief.
\begin{theorem}
\label{th:approx-correction}
Let $\beta$ be a belief such that for any policy, $\pi$, and any starting state, $s$, the total expected reward in any MDP drawn from the belief is between $0$ and $1$; i.e. $ 0 \leq \ex_{s,\pi}^P[V]  \leq 1,$ $\beta$-almost surely. Then the following bound holds for any policy, $\pi$, and any starting state, $s$:
\begin{align*}
&\ex_{\beta} \ex_{s,\pi}^P \sum_{t=0}^{\infty} R_{S_t, A_t} \geq  \ex_{s, \pi}^p \sum_{t=0}^{\infty} \left( \ex_{\beta}[R_{S_t,A_t}] + \sigma^{\beta}_{S_t,A_t}\right)\\
&\text{where}~~\sigma_{s,a}^\beta \deq \sum_{s'} \ex_\beta \left[\min(0,P_{s,a,s'} - \ex_{\beta}[P_{s,a,s'}]) \right].
\end{align*}
\end{theorem}
We first apply Theorem~\ref{th:approx-correction} to the outer term, yielding the following bound:
\begin{align*}
\ex_{s_0,\pi_o}^P \left[\ex_{S_T,\pi_{r}}^P [B_{s_{0}}]\right] = \ex_{s_0,\pi_o}^{\gamma P} \sum_{t=0}^{\infty} (1-\gamma) \left[\ex_{S_t,\pi_{r}}^P [B_{s_{0}}]\right] \\
\geq \ex_{s_0,\pi_{o}}^{\gamma p}\sum_{t=0}^{\infty} \left((1-\gamma)  \ex_{\beta} \ex_{S_t,\pi_r}^P[B_{s_0}] + \gamma \sigma_{S_t,A_t}^\beta\right).
\end{align*}

We, then, apply it again to the inner term:
\begin{align}
\label{eq:inner-bound}
&\ex_{\beta} \ex_{s,\pi_r}^P[B_{s_0}] = \ex_{s,\pi_r}^{P\cdot(1-\ind{s=s_0})} \sum_{t=0}^\infty  1_{S_t = s_0} \geq
\\
&\geq \ex_{s,\pi_r}^{p\cdot (1-\ind{s=s_0})} \sum_{t=0}^\infty \left( \ind{S_t=s_0} + (1- \ind{S_t=s_0}) \sigma_{S_t,A_t}^\beta \right).\notag
\end{align}
Combining the last two results allows us to replace the NP-hard safety constraint with a stricter, but now tractable, constraint. The resulting optimization problem corresponds to the  guaranteed safe, but potentially sub-optimal exploration problem:
\begin{align}
\label{eq:safe-expl-approx}
&\text{maximize }_{\pi_o,\pi_r}\quad\ex_{s_0, \pi_o}^{\gamma p}\sum_{t} \left(r_{S_t, A_t} + \xi^\beta_{S_t, A_t} \right)\\
&\text{s.t.:}~~\ex_{s_0,\pi_{o}}^{\gamma p} \sum_{t=0}^{\infty} \left((1-\gamma) v_{S_t} + \gamma \sigma_{S_t,A_t}^\beta\right) \geq \delta \quad \text{and}\notag\\
&v_{s} = \ex_{s,\pi_r}^{p\cdot (1-\ind{s=s_0})} \sum_{t=0}^\infty \left( \ind{S_t=s_0} + (1- \ind{S_t=s_0}) \sigma_{S_t,A_t}^\beta \right).\notag
\end{align}
The term $v_s$ represents our lower bound for the inner term per Equation~\eqref{eq:inner-bound}, and is simply the value function of the MDP corresponding to the inner term; i.e. the MDP with transition measure $p(1-\ind{s=s_0})$ and reward function $\ind{s=s_0} + (1- \ind{s=s_0}) \sigma_{s,a}^\beta$, under policy $\pi_r$. 
Since the return policy, $\pi_r$, does not appear anywhere else, we can split the safe exploration problem we obtained in Equation~\eqref{eq:safe-expl-approx} into two steps:
\begin{figure*}[t]
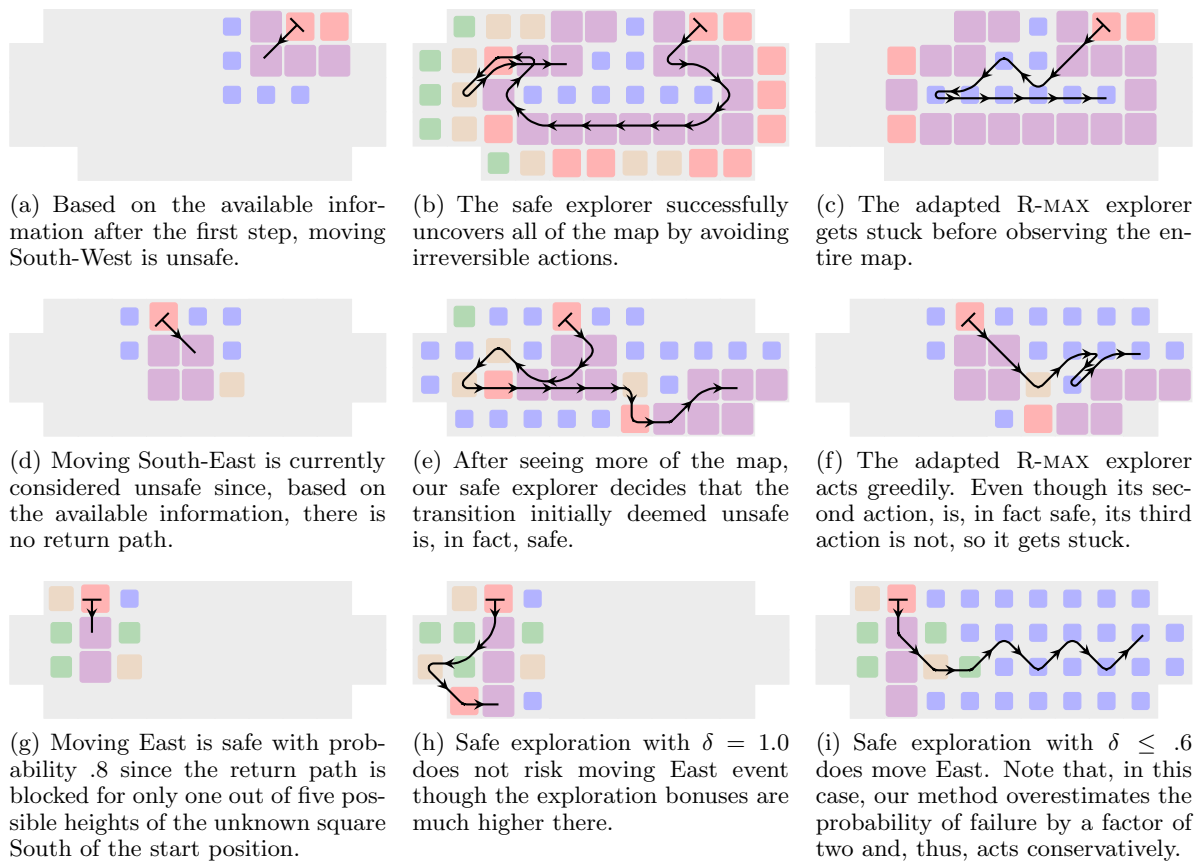

\centering
\subfigure[Based on the available information after the first step, moving South-West is unsafe.]
{\plotpath{./data/expl_movie_tst1_100_safe_adapted_rmax_first.tex}{exa}}~~
\subfigure[The safe explorer successfully uncovers all of the map by avoiding irreversible actions.]
{\plotpath{./data/expl_movie_tst1_100_safe_adapted_rmax.tex}{exb}}~~
\subfigure[The adapted \rmax~explorer gets stuck before observing the entire map. ]
{\plotpath{./data/expl_movie_tst1_adapted_rmax.tex}{exc}}\\
\subfigure[Moving South-East is currently considered unsafe since, based on the available information, there is no return path.]
{\plotpath{./data/expl_movie_tst2_100_safe_adapted_rmax_first.tex}{exd}}~~
\subfigure[After seeing more of the map, our safe explorer decides that the transition initially deemed unsafe is, in fact, safe.]
{\plotpath{./data/expl_movie_tst2_100_safe_adapted_rmax.tex}{exe}}~~
\subfigure[The adapted \rmax~explorer acts greedily. Even though its second action, is, in fact safe, its third action is not, so it gets stuck.]
{\plotpath{./data/expl_movie_tst2_adapted_rmax.tex}{exf}}\\
\subfigure[Moving East is safe with probability $.8$ since the return path is blocked for only one out of five possible heights of the unknown square South of the start position.]
{\plotpath{./data/expl_movie_tst5_100_safe_adapted_rmax_first.tex}{exg}}~~
\subfigure[Safe exploration with $\delta = 1.0$ does not risk moving East event though the exploration bonuses are much higher there.]
{\plotpath{./data/expl_movie_tst5_100_safe_adapted_rmax.tex}{exh}}~~
\subfigure[Safe exploration with $\delta \leq .6$ does move East. Note that, in this case, our method overestimates the probability of failure by a factor of two and, thus, acts conservatively.]
{\plotpath{./data/expl_movie_tst5_80_safe_adapted_rmax.tex}{exi}}
\caption{\label{fig:expl-examples} Exploration experiments in simple grid worlds. See text for full details. Square sizes are proportional to corresponding state heights between 1 and 5. The large, violet squares have a height of 5, while the small, blue squares have a height of 1. Gray spaces represent states that have not yet been observed. Each row corresponds to the same grid world. The first column shows the belief after the first exploration step, while the second and third columns show the entire trajectory followed by different explorers.}
\end{figure*}
\paragraph{Step one:} find the optimal return policy $\pi_r^*$, and corresponding value function $v_{s}^*$, by solving the standard MDP problem below:
\begin{align*}
\ex_{s,\pi_r}^{p\cdot (1-\ind{s=s_0})} \sum_{t=0}^\infty \left( \ind{S_t=s_0} + (1- \ind{S_t=s_0}) \sigma_{S_t,A_t}^\beta \right).
\end{align*}
\paragraph{Step two:} find the optimal exploration policy $\pi_o^*$ under the strict safety constraint, by solving the constrained MDP problem below:
\begin{align*}
&\text{maximize }_{\pi_o}\quad \ex_{s_0, \pi_o}^{\gamma p} \sum_{t} \left(r_{S_t, A_t} + \xi^\beta_{S_t, A_t} \right) \\
&\text{s.t.:}\quad \ex_{s_0,\pi_{o}}^{\gamma p} \sum_{t=0}^{\infty} \left((1-\gamma) v_{S_t}^* + \gamma \sigma_{S_t,A_t}^\beta\right) \geq \delta.
\end{align*}
The first step amounts to solving a standard MDP problem while the second step amounts to solving a constrained MDP problem. As shown by \cite{Altman1999}, both can be solved efficiently either by linear programming, or by value-iteration. In our experiments we used the LP formulation with the state-action occupation measure as optimization variable. Solutions to the constrained MDP problem will usually be stochastic policies, and, in our experiments, we found that following them  sometimes leads to random walks which explore inefficiently. We addressed the issue by de-randomizing the exploration policies in favor of safety. That is, whenever the stochastic policy proposes multiple actions with non-zero measure, we choose the one among them that optimizes the safety objective.

%% file: experiments.tex
\section{Experiments}
\begin{figure*}[t]
\centering
\subfigure[Safe exploration with $\delta=.98$ leads to a model entropy reduction of 7680.
]{\includegraphics[width=.24\textwidth]{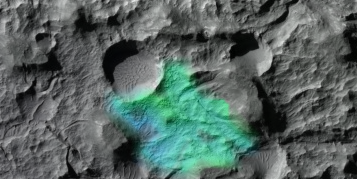}}~
\subfigure[Safe exploration with $\delta=.90$ leads to a model entropy reduction of 12660.
]{\includegraphics[width=.24\textwidth]{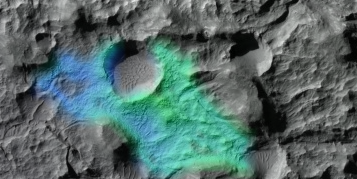}}~
\subfigure[Safe exploration with $\delta = .70$ leads to a model entropy reduction of 35975.
]{\includegraphics[width=.24\textwidth]{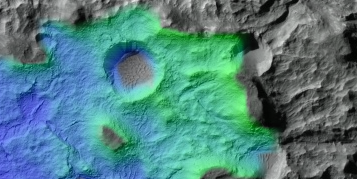}}~
\subfigure[Regular (unsafe, $\delta=0$) exploration leads to a model entropy reduction of 3214.]{\includegraphics[width=.24\textwidth]{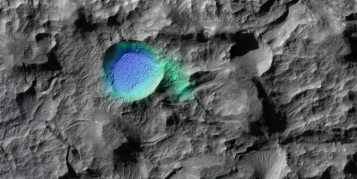}}
\caption{\label{fig:mars-expl} Simulated safe exploration on a 2km by 1km area of Mars at -30.6 degree latitude and 202.2 degrees longitude, for 15000 time steps, at different safety levels. See text for full details. The color saturation is inversely proportional to the standard deviation of the height map under the posterior belief. Full coloration represents a standard deviation of 1cm or less. We report the difference between the entropies of the height model under the prior and the posterior beliefs as a measure of performance. Images: NASA/JPL/University of Arizona.}
\end{figure*}
\subsection{Grid World}
Our first experiment models a terrain exploration problem where the agent has limited sensing capabilities. We consider a simple rectangular grid world, where every state has a height $H_{s}$. From our Bayesian standpoint these heights are independent, uniformly distributed categorical random variables on the set $\{1,2,3,4,5\}$. At any time the agent can attempt to move to any immediately neighboring state. Such move will succeed with probability one if the height of the destination state is no more than one level above the current state; otherwise, the agent remains in the current state with probability one. In other words, the agent can always go down cliffs, but is unable to climb up if they are too steep. Whenever the agent enters a new state it can see the exact heights of all immediately surrounding states. We present this grid world experiment to build intuition and to provide an easily reproducible result. Figure~\ref{fig:expl-examples} shows a number of examples where our exploration method results in intuitively safe behavior, while plain exploration methods lead to clearly unsafe, suboptimal behavior.

Our exploration scheme, which we call \emph{adapted \rmax}, is a modified version of R-max exploration \cite{Brafman2001}, where the exploration bonus of moving between two states is now proportional to the number of neighboring unknown states that would be uncovered as a result of the move, to account for the remote observation model. The safety costs for this exploration setup, as prescribed by Theorem~\ref{th:approx-correction} are:
\begin{align*}
\sigma^\beta_{s,a} = -2 \ex_{\beta}[P_{s,a}](1- \ex_{\beta}[P_{s,a}]) = -2 \var_{\beta}[P_{s,a}]
\end{align*}
where $P_{s,a} \deq \ind{H_{s+a} \leq H_{s} + 1}$ is the probability that attempted move $a$ succeeds in state $s$ and the belief $\beta$ describes the distribution of the heights of unseen states. In practice we found that this correction is a factor of two larger than would be sufficient to give a tight safety bound.

A somewhat counter intuitive result is that adding safety constraints to the exploration objective will, in fact, improve the fraction of squares explored in randomly generated grid worlds. The reason why plain exploration performs so poorly is that the ergodicity assumptions are violated, so efficiency guarantees no longer hold. Figure~6 in the appendix summarizes our exploration performance results. 
\subsection{Martian Terrain}
For our second experiment, we model the problem of autonomously exploring the surface of Mars by a rover such as the \emph{Mars Science Laboratory (MSL)} \cite{Lockwood2006}. The MSL is designed to be remote controlled from Earth but communication suffers a latency of 16.6 minutes. At top speed, it could traverse about 20m before receiving new instructions, so it needs to be able to navigate autonomously. In the future, when such rovers become faster and cheaper to deploy, the ability to plan their paths autonomously will become even more important. The MSL is designed to a static stability of $45$ degrees, but would only be able to climb slopes up to $5$ degrees without slipping \cite{MSLconstraints2007}. Digital terrain models for parts of the surface of Mars are available from the \emph{High Resolution Imaging Science Experiment (HiRISE)} at a scale of 1.00 meter/pixel and accurate to about a quarter of a meter. The MSL would be able to obtain much more accurate terrain models locally by stereo vision.

The state-action space of our model MDP is the same as in the previous experiment, with each state corresponding to a square area of 20 by 20 meters on the surface. We allow only transitions at slopes between -45 and 5 degrees. The heights, $H_s$, are now assumed to be independent Gaussian random variables. Under the prior belief, informed by the HiRISE data, the expected heights and their variances are:
\begin{align*}
\ex_{\beta}[H] &= D^{20}[g \circ h]\quad \text{and}\\ \var_\beta[H] &= D^{20}[g \circ (h - g \circ h)^2] + v_0
\end{align*}
where $h$ are the HiRISE measurements, $g$ is a Gaussian filter with $\sigma = 5$ meters, ``$\circ$'' represents image convolution, $D^{20}$ is the sub-sampling operator and $v_0 = 2^{-4} \text{m}^2$ is our estimate of the variance of HiRISE measurements. We model remote sensing by assuming that the MSL can obtain Gaussian noisy measurements of the height at a distance $d$ away with variance $v(d) = 10^{-6} (d + 1\text{m})^2$. 

To account for this remote sensing model we use a first order approximation of the entropy of $H$ as an exploration bonus:
\begin{align*}
\xi_{s,a}^\beta = \sum_{s'} \var_\beta[H_{s'}]/v(d_{s,s'}).
\end{align*}
Figure~\ref{fig:mars-expl} shows our simulated exploration results for a $2$km by $1$km area at $-30.6$ degrees latitude and $202.2$ degrees longitude \cite{PSP_010228_1490}. Safe exploration at level $1.0$ is no longer possible, but, even at a conservative safety level of $.98$, our method covers more ground than the regular (unsafe) exploration method which promptly get stuck in a crater. Imposing the safety constraint naively, with respect to the expected transition measure, as argued against at the end of Section~\ref{sec:objective-counter}, performs as poorly as unsafe exploration even if the constraint is set at $.98$.
\subsection{Computation Time}
\begin{table}[t]
\caption{Per-step planning times for the $50\times 100$ grid world used in the Mars exploration experiments, with $\gamma=.999$.}
\label{table:running-time}
\vskip 0.15in
\begin{center}
\begin{small}
\begin{tabular}{lr}
\hline
\abovespace\belowspace
Problem setting & Planning time (s) \\
\hline
\abovespace
Safe exploration at .98 & $5.86 \pm 1.47$  \\
Safe exploration at .90 & $10.94 \pm 7.14$ \\
Safe exploration at .70 & $4.57 \pm 3.19$ \\
Naive constraint at .98 & $2.55 \pm 0.42$ \\
\belowspace
Regular (unsafe) exploration & $1.62 \pm 0.26$  \\
\hline
\end{tabular}
\end{small}
\end{center}
\vskip -0.1in
\end{table}
We implemented our algorithm in Python 2.7.2.7, using Numpy 1.5.1 for dense array manipulation, SciPy 0.9.0 for sparse matrix manipulation and Mosek 6.0.0.119 for linear programming. The discount factor was set to $.99$ for the grid world experiment and to $.999$ for Mars exploration. In the latter experiment we also restricted precision to $10^{-6}$ to avoid numerical instabilities in the LP solver. Table~\ref{table:running-time} summarizes planning times for our Mars exploration experiments.

%% file: discussion.tex
\section{Discussion}
In addition to the safety formulation we discussed in Section~\ref{sec:safety-constraint}, out framework also supports a number of other safety criteria that we did not discuss due to space constraints:
\begin{itemize}[noitemsep,nolistsep]
\item Stricter ergodicity ensuring that return is possible within some horizon, $H$, not just eventually, with probability $\delta$.
\item Ensuring that the probability of leaving some pre-defined safe set of state-actions is lower than $1-\delta$.
\item Ensuring that the expected total reward under the belief is higher than $\delta$.
\end{itemize}
Additionally, any number and combination of these constraints at different $\delta$-levels can be imposed simultaneously.

\section*{Acknowledgements}
This material is based upon work supported in part by NSF under award IIS-0931463, by ARO under the MAST program, by a Sloan Fellowship, by a gift from
Intel, by the U. S. Army Research Laboratory and the U. S. Army Research Office under contract/grant number W911NF-11-1-0391.

%% file: appendix.tex
\section*{Appendix}
\begin{figure*}[t]
\centering
\plotexplsucc{expl_succ}
\caption{\label{fig:expl-succ} Exploration efficiency comparison. We are showing the median, the upper and the lower quartiles of the fraction of the grid world that was uncovered by different explorers in randomly generated grid worlds. The ``amount'' of non-ergodicity is controlled by randomly making a fraction of the squares inaccessible (walls). We ran 1000, 500, 100,20 experiments for grids of sizes $10^2$, $20^2$, $30^2$ and $40^2$ respectively. We are comparing against our own adapted \rmax~exploration objective, the original \rmax~objective \cite{Brafman2001} and the Near-Bayesian exploration objective \cite{Kolter2009}. The last two behave identically in our grid world environment, since, once a state is visited, all transitions out of that state are precisely revealed.}
\end{figure*}
\subsection*{Proof of Theorem \ref{th:np-hard}.}
\begin{figure}[tb]
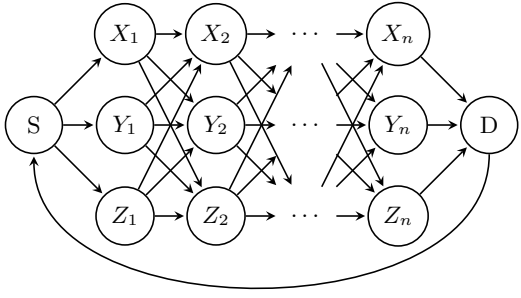

\centering
\nphardproofdiagram{np_hard_proof_diagram}
\caption{\label{fig:nphardproofdiagram} MDP reduction of the 3SAT problem.}
\end{figure}
\begin{proof}
We will prove the theorem by reducing the satisfiability problem in conjunctive normal form with three variables (3SAT) to the problem of deciding whether there exists a $p$-safe policy for a belief that we will construct. The 3SAT problem amounts to deciding whether there exists an assignment to boolean variables $\{U_k\}$ such that the following expression is true:
\begin{align*}
(X_1 \vee Y_1 \vee Z_1) \wedge \cdots \wedge (X_n \vee Y_n \vee Z_n)
\end{align*}
where each of the variables $X_i, Y_i, Z_i$ equals one of the variables in $\{U_k\}$, possibly negated.

We start by constructing an MDP to represent this problem as shown in Figure~\ref{fig:nphardproofdiagram}. In addition to actions corresponding to the outgoing arrows, the agent also has the option of remaining in the same state. A transition from some state to another state will succeed if and only if the boolean variable corresponding to the origin state is true. The boolean variable associated to states $S$ and $D$ are always true. Our belief is the uniform distribution over truth values of the boolean variables $\{U_k\}$.

Now consider the following simple policy: from $D$ go to $S$ and then stay in $S$. For any recall time $T>0$, the recall event will find the agent in state $S$, so the policy is $p$-safe for any $p>0$ if and only if the belief assigns a non-zero measure to MDPs in which state $D$ is accessible form state $S$, so if any only if there exists at least one boolean assignment for the $\{U_k\}$ such that state $D$ is accessible from $S$. It is easy to see that, this is the case if and only if the 3SAT formula is satisfied, and this observation completes the reduction. 

This result should come as no surprise since similar optimization problems have been shown to be NP-hard in the context of \emph{Partially Observable Markov Decision Processes} \cite{Blondel2000}.

\end{proof}
\subsection*{Proof of Theorem \ref{th:approx-correction}.}
\begin{proof}
The result is an immediate consequence of the following Lemma.
\end{proof}
\begin{lemma}
\label{th:exact-correction}
Given a belief $\beta$ and a policy $\pi$, there exists a policy dependent reward correction, $\sigma^{\beta, \pi}$, defined below, such that the MDP with transition measure $p \deq \ex_{\beta} P$ and rewards $r + \sigma^{\beta, \pi}$, where $r \deq \ex_{\beta} R$, has the same expected total return as the belief for any initial distribution. Formally:
\begin{align*}
\forall \rho \quad \ex_{\beta} \ex_{\rho,\pi}^P \sum_{t=0}^{\infty} R_{S_t,A_t} =  \ex_{\rho,\pi}^p \sum_{t=0}^{\infty} \left( r_{s,a} + \sigma^{\beta,\pi}_{s,a}\right)\\
\sigma^{\beta,\pi}_{s,a} \deq \sum_{s'} \ex_{\beta} \left[ (P_{s,a,s'} - \ex_{\beta}[P_{s,a,s'}]) \ex_{s',\pi,P}[V] \right].
\end{align*}
\begin{proof}
The Markov property under belief $\beta$ reads:
\begin{align*}
\ex_\beta\ex_{s,\pi}^P[V] &= \sum_{a} \pi_{s,a} \ex_{\beta}[R_{s,a}] +\\
&+ \sum_{a}\pi_{s,a}\sum_{s'} \ex_{\beta}[P_{s,a,s'} \ex_{s',\pi}^P[V]].
\end{align*}
The Markov property assuming expected transition frequencies and expected rewards with safety penalty is:
\begin{align*}
\ex_{s,\pi}^p[\bar V] &= \sum_{a} \pi_{s,a} (r_{s,a} + \sigma^{\beta,\pi}_{s,a})+\\
&+ \sum_{a}\pi_{s,a}\sum_{s'} p_{s,a,s'} \ex_{s',\pi}^p [\bar V]].
\end{align*}
Now let $\Delta_{s} \deq \ex_\beta \ex_{s,\pi}^P[V] - \ex_{s,\pi}^p[\bar V]$. By subtracting the first two equations we get that:
\begin{align*}
\Delta_s = \sum_{a}\pi_{s,a} \sum_{s'} p_{s,a,s'} \Delta_{s'}.
\end{align*}
We can see that $\Delta$ satisfies the same equation as the value function in an MDP with transition measure $p$ and zero rewards. Since the value function in such an MDP is uniquely defined and identically zero, we conclude $\Delta_s=0$.
\end{proof}
\end{lemma}